\documentclass[letterpaper, 10 pt, conference]{ieeeconf}  

\IEEEoverridecommandlockouts                              

\usepackage[utf8]{inputenc}
\usepackage[top=55pt, left=54pt, right=54pt, bottom=54pt]{geometry}

\usepackage{xcolor}
\usepackage{graphicx}
\usepackage{subcaption}
\usepackage{lipsum}  
\usepackage{balance}
\usepackage{rotating}
\usepackage{transparent}
\usepackage[font=small]{caption}

\usepackage{array}
\usepackage{tabularx}
\usepackage{makecell}
\usepackage{multirow}
\usepackage{setspace}
\usepackage{booktabs}

\makeatletter
\let\NAT@parse\undefined
\makeatother          
\usepackage{hyperref}
\usepackage{standalone}

\usepackage[nameinlink,capitalise]{cleveref}
\crefformat{footnote}{#2\footnotemark[#1]#3}

\usepackage{tikz}
\usetikzlibrary{patterns,positioning,arrows,arrows.meta,calc,shapes,pgfplots.groupplots,fit,backgrounds}

\title{\LARGE \bf
Happily Error After: Framework Development and User Study for Correcting Robot Perception Errors in Virtual Reality 
}

\author{Maciej K. Wozniak*$^{1}$, Rebecca Stower*$^{1}$, Patric Jensfelt$^{1}$, and André Pereira$^{2}$
\thanks{*Both authors contributed equally to this work}
\thanks{$^{1}$Maciej K. Wozniak, Rebecca Stower and Patric Jensfelt are with the Division of Robotics, Perception, and Learning, KTH Royal Institute of Technology, Stockholm, Sweden}
\thanks{$^{2}$André Pereira is with the Division of Speech, Music, and Hearing, KTH Royal Institute of Technology, Stockholm, Sweden}%
}

\begin{document}

\maketitle
\thispagestyle{empty}
\pagestyle{empty}

\begin{abstract}
While we can see robots in more areas of our lives, they still make errors. One common cause of failure stems from the robot perception module when detecting objects. Allowing users to correct such errors can help improve the interaction and prevent the same errors in the future. Consequently, we investigate the effectiveness of a virtual reality (VR) framework for correcting perception errors of a Franka Panda robot. We conducted a user study with 56 participants who interacted with the robot using both VR and screen interfaces. Participants learned to collaborate with the robot faster in the VR interface compared to the screen interface. Additionally, participants found the VR interface more immersive, enjoyable, and expressed a preference for using it again. These findings suggest that VR interfaces may offer advantages over screen interfaces for human-robot interaction in erroneous environments.
\end{abstract}

\maketitle
\thispagestyle{empty}
\pagestyle{empty}

\section{Introduction}
\label{sec:intro}

Virtual, Augmented and Mixed reality technologies (VAM) have been used in various fields to enhance user experience and provide immersive and realistic environments \cite{roldan2017multi}. In recent years, VAM has been applied in different robotics projects \cite{wozniak2023virtual}. In particular, human-robot interaction (HRI) can often benefit from additional devices to complement the robot's capabilities. VAM and screen-based interfaces each offer different advantages and disadvantages. While traditional screens are widely available and accessible, with VAM, users can interact with robots in a virtual environment that more closely resembles the real world, allowing for more natural and intuitive interactions. However, for some tasks, traditional screen-based interfaces may be more convenient and efficient as VAM users need specialized equipment, which may limit its widespread adoption. It is not always trivial to decide which technology is better when it comes to HRI applications. Despite this, there is a lack of studies that compare the effectiveness of VAM and screen-based interfaces for HRI. 




While robots are increasingly being used in a variety of applications, from manufacturing and healthcare to household and entertainment, they still make errors. Oftentimes, these errors are due to imperfections in their perception systems~\cite{schutte2017robot}. These errors can lead to incorrect decisions and actions by the robot, which can have serious consequences. One possible way of recovering from perception failures is by allowing users to correct the error. Users often have knowledge about specific tasks, which can help improve the accuracy of the robot's perception system. Moreover, giving users more autonomy and control can also improve their task performance and perception of the interaction \cite{wallkotter2021}, which is essential for successful long-term human-robot collaboration.


\begin{figure}
     \centering
     \resizebox{\columnwidth}{!}{%
     \begin{subfigure}[t]{0.345\columnwidth}
         \centering
         \includegraphics[width=\textwidth]{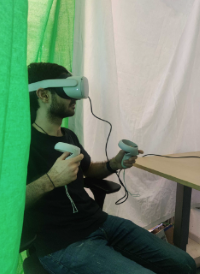}
         \caption{Participant in the VR}
         \label{fig:collJ}
     \end{subfigure}
     \hfill
     \begin{subfigure}[t]{0.62\columnwidth}
         \centering
         \includegraphics[width=\textwidth]{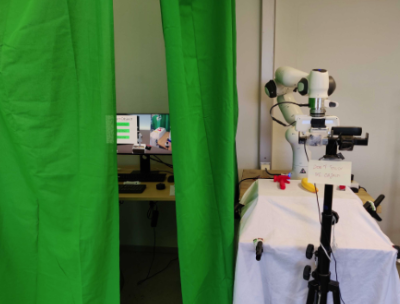}
         \caption{Robot together with the screen interface}
         \label{fig:collrobenv}
     \end{subfigure}
     }
     \par\smallskip 
     \resizebox{\columnwidth}{!}{%

     \begin{subfigure}[t]{0.293\columnwidth}
         \centering
         \includegraphics[width=\textwidth]{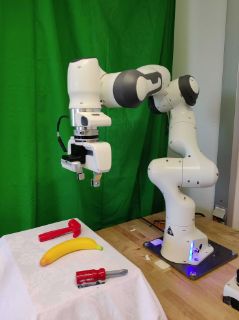}
         \caption{Objects}
         \label{fig:collobj}
     \end{subfigure}
     \hfill
     \begin{subfigure}[t]{0.66\columnwidth}
         \centering
         \includegraphics[width=\textwidth]{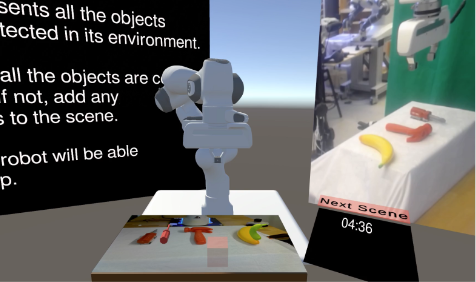}
         \caption{User interface}
         \label{fig:collui}
     \end{subfigure}
     }
     \setlength{\belowcaptionskip}{-15pt}
        \caption{Experimental set-up for the User Study.}
        \label{fig:collage}
\end{figure}


In this work, we investigate how effective Virtual Reality (VR) frameworks are for correcting robot errors. We compare a VR interface with a 2D screen interface. We chose VR because it provides an immersive remote interaction with the robot. We conducted a user study where 56 participants used both interfaces to correct a robot's perception errors. We measured both task performance and participants' subjective perceptions of each interface. 

There are two main contributions of this paper. First, we design and implement a VR framework for interacting with the robot and correcting its perception errors. Second, we conduct a user study assessing how well the VR framework performs in comparison to a screen-based interface. The code, hypotheses, statistical data, and other supplementary materials are all publicly available\footnote{\url{https://github.com/maxiuw/pickandplace}\label{foot:projectlink}}.


\section{Related Work}

\subsection{VAM User Interfaces for Human-Robot Interaction}
One of the most common purposes of VAM in HRI is teleoperation \cite{rosen2018testing}. In this context, VAM has shown to be more immersive and often more intuitive for operators from various fields than baseline solutions (such as industrial controllers or gamepads). VAM systems can also improve human understanding of robot behaviours, thereby improving the quality of the interaction \cite{walker2018communicating}. The purpose of VAM  varies from projecting safe or no-go zones onto the factory floor \cite{chandan2021arroch}, to the ability to visualise and modify the robot's trajectory \cite{wozniakvirtual}, or even visualising sensor readings in virtual reality~\cite{groechel2022reimagining}. Mutual human-robot understanding can also be improved by visualizing the future state of the robots \cite{chandan2021arroch}. VAM has also found useful applications in industrial settings, focusing on factory workers' training with a digital twin~\cite{wang2021interactive}, human-robot collaboration~\cite{ortenzi2022robot}, assembly processes~\cite{guzzi2022conv}, or teaching a robot new tasks by demonstration \cite{9561844}.

VAM also has different applications in HRI. Hedayati et al. \cite{hedayati2018improving} demonstrate that Augmented Reality (AR) is superior to other existing systems when it comes to users' attention and performance. Groechel et al. \cite{groechel2019using} also found that Mixed Reality (MR) helps users to better understand robots' intentions and gestures. Others suggest that VAM can elevate users' feeling of presence \cite{pereira2017augmented} or reduce the complexity of the task \cite{ikeda2021ar}. Additionally, Head-Mounted Displays (HMD) are equipped with different sensors enabling eye or hand tracking, which helps measure user engagement level or facilitate robot teleoperation \cite{rosen2020mixed}. Virtual environments can also enable VAM-HRI studies without the need for a physical robot \cite{groechel2022reimagining}. 

Nevertheless, none of this work considers interaction with the robot and correcting its perceptual errors, as we do in our framework. In this work, we therefore focus on how humans can leverage VR to correct robots' misunderstanding of the environment, caused by perception module failures.



\subsection{Robot Failures}
Several taxonomies and classification systems have been developed relating to the identification and resolution of robot failures in HRI \cite{tolmeijer2020taxonomy, brooks2016analysis}. In our work, we investigate technical/system failures, which occur when the system fails to do what it was designed for \cite{tolmeijer2020taxonomy}.  Although technical failures have received some attention with humanoid robots, comparatively little work has been done investigating how failures of robotic arms are perceived.

Other works relating to robot failures and recovery have investigated the role of explainability, with the idea that providing explanations behind (faulty) robot behaviour can help mitigate the impacts of said failures \cite{esterwood2022literature}. Nonetheless, most error recovery strategies to date focus on actions that the robot can take to resolve a failure. In contrast, there is limited work looking at how humans can proactively react to robot errors to prevent task breakdowns. 

To address these gaps, we propose a VR framework where users are able to directly correct robot perceptual errors. By using VR as an interface through which users can interact with the real-world robot, we aim to both, increase the transparency of the interaction by allowing users to view the robots' perception of the world in real-time, as well as give users a greater sense of autonomy and control during the interaction. We have so far come across very few studies in HRI which have directly compared VR and screen interfaces \cite{mara2021user}. However, such comparisons are useful, as they can help determine the benefits of VAM interfaces over current and more commercially available technologies, as well as inform the future design of such interfaces.






\section{Framework and Task Design}
\label{sec:tech}
The task designed for this study is composed of four steps. Participants need to (i) assess how the robot perceives the real world environment (using either screen or VR), (ii) correct the robot's understanding of the environment in the virtual world, (iii) verify a proposed action of the virtual robot, and (iv) deploy the same movement on the real world robot.


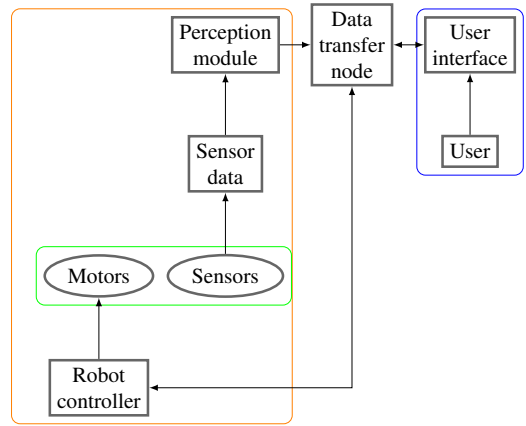
\begin{figure}[b]
    \centering
    \resizebox{0.8\columnwidth}{!}{%
         \begin{tikzpicture}[squarednode/.style={rectangle, draw=black!60,  very thick, minimum size=5mm},every text node part/.style={align=center},   dot/.style={
        circle,
        minimum size=3pt}, ellipsenode/.style={ellipse, draw=black!60,  very thick, minimum size=5mm}]
    \node[squarednode]      (realworld)                              {Sensor \\ data};
    \node[ellipsenode]        (sensors)       [below=of realworld] {Sensors};

    \node[ellipsenode]        (motors)       [left= 0.2cm of sensors] {Motors};
    \node[squarednode]        (DNN)       [above=of realworld] {Perception \\module};
    \node[squarednode]        (tcp)       [right= 0.5cm of DNN] {Data\\transfer\\node};
    \node[squarednode]        (vr)       [right = 0.5cm of tcp] {User  \\ interface};
    \node[squarednode]        (user)       [below=of vr] {User};
    \node[dot] (e) at (-3.2,1.5) {};
    \node[dot] (f) at (0.8,-3) {};

    \node[squarednode]        (motion)       [below=of motors] {Robot \\ controller};
    \node (box) [draw=orange,rounded corners,fit = (realworld) (DNN)  (motion) (e) (f)] {};
    \node (box) [draw=blue,rounded corners,fit = (user) (vr)] {};
    \node (box) [draw=green,rounded corners,fit = (sensors) (motors)] {};
    
    \draw[-latex] (sensors.north) -- (realworld.south);
    \draw[-latex] (realworld.north) -- (DNN.south);
    \draw[-latex] (DNN.east) -- (tcp.west);
    \draw[latex-latex] (tcp.east) -- (vr.west);
    \draw[-latex] (user.north) -- (vr.south);
    \draw[latex-latex] (tcp.south) |- (motion.east);
    \draw[-latex] (motion.north) -| (motors.south);


\end{tikzpicture}
    }
        \caption{Experimental setup. The elements within the \textcolor{orange}{orange box} correspond to nodes and parts connected with the robot's motion and perception, whereas the ones in the \textcolor{green}{green box} are explicitly corresponding to its hardware. Nodes inside the \textcolor{blue}{blue box} are connected to the user and screen or VR interface.}
        \label{fig:pipeline}
        
\end{figure}

Our proposed framework integrates several systems; the cameras and associated perception module, and the real-world robot. It has been tested with the Franka Panda~\cite{wozniak2023you}, Niryo One~\cite{wozniakvirtual} robotic arms, and the Baxter humanoid robot~\cite{moletta2023virtual}. The framework can also be adapted to work with other robots. While in this work we focused exclusively on correcting perception errors, more functions of the framework are described in~\cite{wozniak2023you,wozniakvirtual}. A recording of the full interaction can be found in the supplementary video\cref{foot:projectlink}.

Planning is done using the MoveIt library and Robot Operating System (ROS) is used to transfer data and run different tasks on separate nodes. Our interface is developed and tested using Unity (2020.3 version) and the Oculus Quest 2 headset. In addition to the VR interface, we also designed a screen interface with a mouse and keyboard. We adapted the same application to both VR and screen interfaces, allowing us to compare the two whilst ensuring that users could execute the same actions with the different controllers (see \cref{subsec:control}).

When the user interacts with the robot through the VR headset, planning and object detection must be run on separate machines, which creates the challenge of efficiently transferring data between different nodes and the VR headset. To address this challenge, we only transfer the output of the detection network between the robot and the virtual environment to facilitate information exchange, improve the user experience, and minimise the necessary bandwidth. 


\subsection{Perception Module}
\label{subsec:perception}

The main camera streams a video that is then processed by a deep learning model~\cite{wu2019detectron2} to segment the scene and detect objects. The resulting segmentation masks and bounding boxes provide information about the location and class of detected objects, which is used to accurately position and add objects to the VR scene. The perception module generates the VR environment based on this information, allowing the user to interact with the detected objects using the robot. 

In the virtual world, participants can also see a real-time camera image, projected on the virtual table where the objects are located (hereon referred to as \textit{passthrough}). The passthrough view allows the user to see where the objects are in the real world and determine which object is missing.

\subsection{Correcting Perception Errors}
Detected objects appear in the virtual world as virtual representations of those objects. In addition to the passthrough view, there is also a live-stream video of the real-world robot projected in the UI. Users can check both camera views to verify if all objects have been successfully detected. If the robot does not detect an object, users can use an \textit{add object} menu to add the missing object to the virtual environment. Now, users can interact with these objects and they are also considered for collision avoidance in the robot's trajectory planner.


Next, they can use an \textit{action} menu to tell the virtual robot to pick up the virtual object that was just added (\textit{pick up object}). Finally, once the virtual robot has finished demonstrating the movement, they can send the same action to the real-world robot (\textit{move real robot}). At the end of a trial, both the real-world and virtual robots are reset to their original positions.

\subsection{Controls}
\label{subsec:control}


Users were able to interact with the robot through different controllers, depending on the type of graphical user interface they were using. In the VR interface, users had two controllers: left and right. On the left controller, there were two active buttons: \textit{X} and \textit{Y}. The \textit{X} button brings up the \textit{add object} menu, with a list of all available objects. The \textit{Y} button brings up the \textit{action} menu with two actions; pick up object, which  tells the virtual robot to pick up a virtual object, and move real robot, which sends the same movement that the virtual robot just performed to the real-world robot. The right-hand controller works as a pointer. The users had to point at the menu and press the \textit{trigger} button to choose the option from the menu or point it at the object and press the \textit{grip} button to pick it up.

Similarly, in the screen interface, the \textit{X} and \textit{Y} buttons on the keyboard  bring up the same menus. The mouse cursor functions as a pointer, whilst the \textit{left} mouse button could be used to select menu options or pick up an object. 

Users could move around the environment in both interfaces. In the VR interface, the virtual view followed the user's head rotation and position. In the screen interface, users could use the WSAD keys to move up, down, left and right. They could also rotate the view by holding the \textit{right} mouse button and moving the mouse, as well as use the scroll wheel to zoom in and out. To return to their initial position, users could press the \textit{Oculus} button on the VR controller, or space bar on the keyboard. 



%



\section{User Study}

\subsection{Participants}
Calls for participation were disseminated via word of mouth, flyers and social media networks. 59 participants were recruited, 3 of which were excluded due to language barriers or failing to understand the task, leaving a final sample of 56 ($M_{age} = 26.45, SD = 4.78$). 23 participants identified as female, 33 as male, and none of any other gender. 29 participants reported using corrective glasses or contact lenses. Before beginning the experiment, participants were given an information sheet and asked to indicate their consent to participate. The experiment took approximately 45 minutes, and participants received a 100SEK voucher as thanks.


\subsection{Design and Procedure}
Upon arriving, participants were informed that they would perform a task with a real-life robot, during which they would use two different interfaces (VR, screen) to interact with the robot. They then completed a short questionnaire with demographic information (age, gender, handedness, and if they required corrective glasses or contact lenses) and questions assessing their familiarity with VAM systems and with robots (FAM-VAM and FAM-Robot). 

We employed a one-way within groups design, with the type of interface used to correct robot errors (VR, screen) manipulated as the independent variable. Participants were randomly assigned to either start with the VR or screen interface. They were then given instructions specific to that interface for how to complete the error-correction task. For each interface, participants completed 4 trials. The first trial was used as a training phase for participants to familiarise themselves with the task and controls. 




In front of the physical Franka Panda robot, there was a table with 3 toy objects: a hammer, screwdriver, and banana. The placement of the objects on the table was random, such that each participant experienced a different object configuration (e.g., banana left, screwdriver middle, and hammer right). 

Each trial consisted of the robot failing to detect one of the objects. Although in real-world scenarios many different kinds of errors are possible, these might not be consistent. We therefore operationalised failure as an object detection error due to the perception module. To ensure that each trial includes one missing object, the real-world robot was purposely made to fail to detect one of the 3 objects. That is, only 2 out of the 3 real-world objects had corresponding virtual representations. Which object was missing was pseudo-randomised, with each object missing at least once (as we had 4 trials in total, one object was missing twice). After 4 trials with one interface, participants were asked to complete a series of questionnaires about that specific interface (see \cref{sec:measures}). They then switched to the second interface (with corresponding instructions), and the same procedure was repeated. 







\subsection{Measures}
\label{sec:measures}
\subsubsection{Objective}
As measures of task performance, we extracted the time (in seconds) from the beginning of a trial until the missing object was added (\textit{time to add}) and the time taken from adding the object until asking the virtual robot to pick it up (\textit{time to place}). Time to add the object was intended as a measure of how easy it was to determine which object is missing, whereas time to place captures how easy it was to manipulate the object. We did not record total trial time, as this was dependent on the trajectory planned by the robot and how long it took to execute the movement in the virtual and real-world. 

We also measured the distance of the virtual object from the real-life object (\textit{final distance}). This was calculated as the distance (in meters) from the center point of the bounding box capturing the real-life object to the center point of the virtual object placed by the participant. Finally, we counted the number of times per trial participants added the missing object to the virtual environment (\textit{n-missing}).


\subsubsection{Subjective} 
We measured participants' perceptions of each UI using a series of self-report scales targeting presence (adapted from \cite{witmer1998measuring}) and enjoyment (using items from \cite{diefenbach2014hedonic}). We modified some items to better fit our study context and to be consistent between scales and conditions (e.g., switching ``virtual reality interface'' and ``screen and keyboard interface'' between conditions). We converted all answer scales to a 5-point Likert scale ranging from 1 - \textit{Strongly Disagree} to 5 - \textit{Strongly Agree}. 

After participants had interacted with both UI's, they were asked 3 \textit{forced-choice} questions, adapted from the Technology Acceptance Model \cite{venkatesh2012consumer}, about which interface  (VR, screen) they found easier to use, more useful, and would prefer to interact with again. Finally, we gave participants a free-text box where they could explain the reasoning behind their choices, as well as any other comments.


\section{Hypotheses}
All hypotheses were preregistered on the Open Science Framework\cref{foot:projectlink}.

\begin{enumerate}
    \item The VR interface will be preferred over the screen interface in terms of: 
    \begin{itemize}
        \item[a.] Usability
        \item[b.] Ease of Use.
        \item[c.] Future use intentions.
     \end{itemize}
    \item Participants will have better task performance with the VR system:
    \begin{itemize}
        \item[a.] Participants will have more accurate object placement when adding virtual objects to the environment.
        \item[b.] There will be an interaction between condition and trial number on the time taken to complete trials, such that the VR condition will have longer initial times and then decrease in subsequent trials, whereas the screen will have similar times across all trials.
    \end{itemize}
    \item Subjective ratings on the self–report questionnaires will be higher for the VR system compared to the screen interface for:
        \begin{itemize}
            \item[a.] Enjoyment.
            \item[b.] Presence.
        \end{itemize}
\end{enumerate}


\section{Results}
All analyses were performed in R version 4.2.2. We separated the task performance data into the \textit{training phase} (Trial 0) and \textit{experimental phase} (Trials 1, 2 and 3). In all figures, \textit{ns} indicates non-significance, * indicates significance at the $p < .05$ level, ** at the $p < .01$ level and *** at the $p < .001$ level.

\subsection{Training Phase}
For the four metrics of task performance (time to add, time to place, final distance, and \textit{n}-missing) we first examined the means from the training phase, see \cref{tab:training_means}. 

\begin{table}[b]
\scriptsize
\caption{Means and Standard Deviations for Each Measure of Task Performance in Each Condition}
\label{tab:training_means}
\begin{tabular}{llllll}
\toprule
                & \makecell{\textbf{Time to}\\\textbf{Add (s)}} & \makecell{\textbf{Time to}\\\textbf{Place (s)}} & \makecell{\textbf{Final}\\\textbf{Distance (m)}} & \makecell{\textbf{\textit{n}-missing}} \\ \midrule
\textbf{VR}     & 80.69 (52.22)                   & 135.24 (70.00)                    & 0.05 (0.07)      & 1.15 (0.40)         \\
\textbf{Screen} & 35.36 (24.50)                    & 114.01 (61.72)                    & 0.04 (0.05)       & 3.07 (1.69)          \\ \bottomrule
\end{tabular}
\end{table}

Wilcoxon signed rank tests comparing each condition suggest that in the training trial, the VR took significantly longer to add the object, $V = 223, p <.001$, however, the missing object was added significantly less times $V = 982.5, p <.001$. There was no difference in time taken to place the object $V = 550, p = .065$ or in the final distance of the object $V = 657, p = .608$. The difference in the time to add the object is likely due to participants requiring more time to familiarise themselves with the VR controllers. 

\subsection{Experimental Phase}
We fitted linear mixed models (LMM) with identity functions for the time taken to add and place the object in the virtual environment. We corrected for non-normal distributions of residuals using Box-Cox transformations.

Each LMM included fixed effects of condition (VR, Screen), order (VR First, Screen First), trial (1, 2, 3), familiarity with VAM systems (FAM-VAM) and familiarity with robots (FAM-Robots). We mean-centred the values for familiarity with VAM/Robots.  We also included a random effect of trial nested within participant ID. \textit{F}-values are reported where it is not possible to obtain a single slope estimate (i.e., for categorical variables with $>2$ levels). 



\subsubsection{Time to Add Object}
The model for the time taken by participants to add the missing object to the virtual environment (Model 1) showed significant fixed effects of the order in which participants interacted with each UI, and the trial number, see \cref{tab:add_fixef}.

\begin{table}[b]
    \centering
    \footnotesize
    \caption{Fixed Effects for Time to Add Missing Object}
    \resizebox{\columnwidth}{!}{%
    \label{tab:add_fixef}
    \begin{tabular}{lllllll}
      \toprule
    \multirow{2}{*}{\textbf{Variable}} & \multicolumn{6}{c}{\textbf{Model 1}} \\ \cmidrule(lr){2-7} 
     & \textit{b} & \textit{se} & \textit{t} & \textit{F} & \textit{p} & \textit{95\% CIs} \\ \midrule
    (Intercept) & 1.70 & 0.05 & 33.34 & & $<.001$ & [1.60, 1.80 ]  \\ 
      Condition & -0.02 & 0.04 & -0.54 & & .585 & [-0.10, 0.06]  \\ 
      Order & -0.13 & 0.05 & -2.40 & & .016 & [-0.23, -0.02]  \\
      FAM-VAM & -0.02 & 0.04 & -0.57 & & .561 & [-0.09, 0.05] \\
      FAM-Robot & -0.05 & 0.03 & -1.79 & & .066 & [-0.1, 0.00]  \\
      Trial &  &  &  & 12.42 & .002 \\ \midrule
      AIC & 304.24 & & & & \\ \bottomrule
    \end{tabular}%
    }
    \end{table}

\begin{figure}[b]
    \centering
    \includegraphics[width=0.5\textwidth]{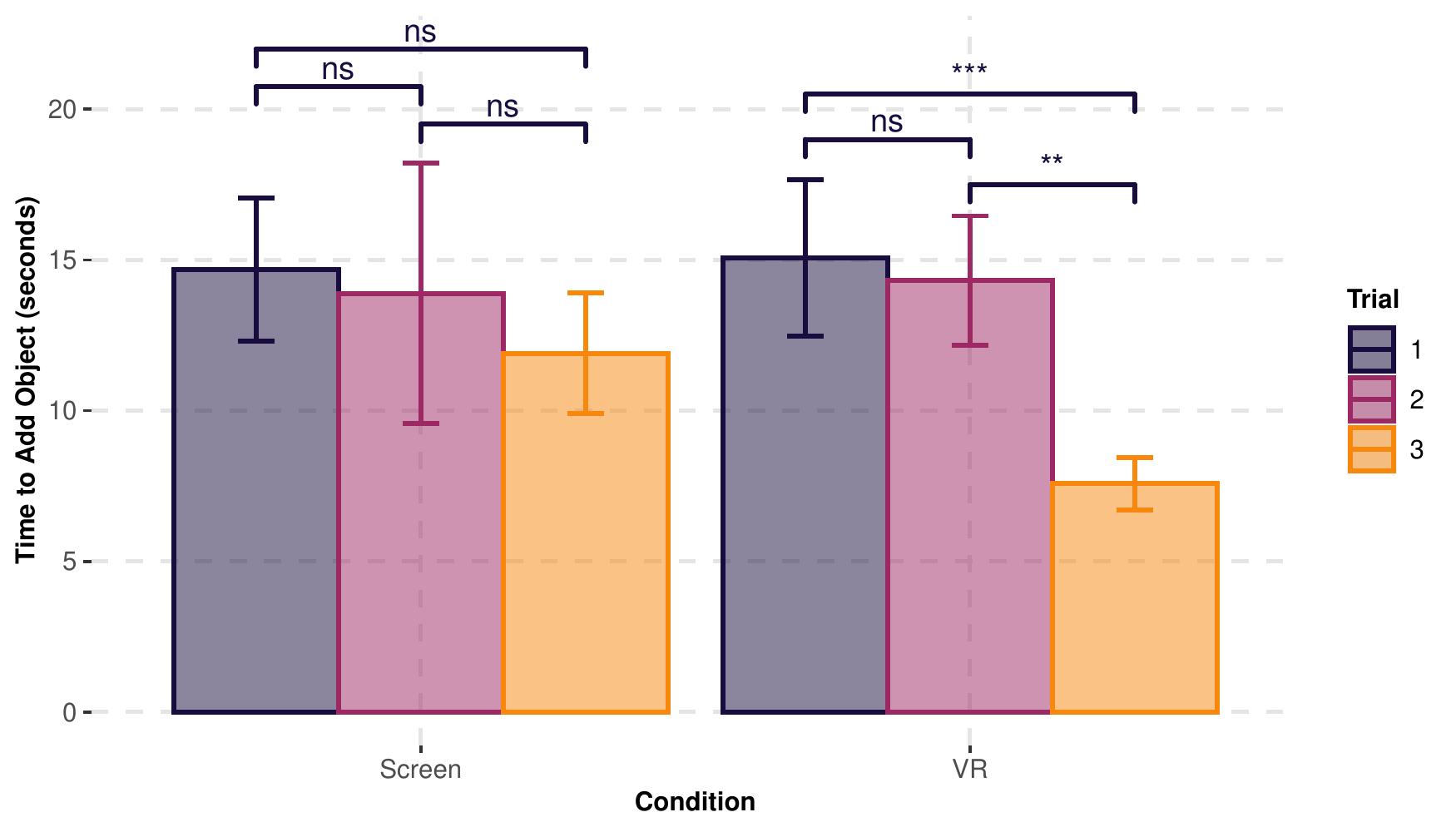}
    \caption{Interaction between Condition and Trial for Time to Add Object.}
    \label{fig:add_interact}
\end{figure}

We then fitted two additional models (Model 2 and Model 3) with the interaction effects between condition, order, and trial, see \cref{tab:add_interact}. Comparing the three models suggests that Model 2 provides the best fit for the data, with a significant condition by trial interaction (see \cref{fig:add_interact}), as well as a significant condition by order interaction. 

\begin{table}[h!]
    \centering
    \footnotesize
    \caption{Interaction Effects for Time to Add Missing Object}
    \label{tab:add_interact}
    \begin{tabular}{lllll}
    \toprule
                          & \multicolumn{2}{c}{\textbf{Model 2}} &  \multicolumn{2}{c}{\textbf{Model 3}}  \\ \cmidrule(lr){2-3} \cmidrule(lr){4-5} 
                          & \textit{F}        & \textit{p} &  \textit{F}       & \textit{p} \\ \midrule
      Condition x Trial   & 9.03 & .011 & & \\
      Condition x Order   &  23.24 & $<.001$  & & \\
      Order x Trial       &  0.81 & .668  & & \\
      Condition x Trial x Order &  &  & 35.89 & $<.001$ \\ \addlinespace \midrule
      AIC                  & 281.98 & & 282.35 & \\ \bottomrule
    \end{tabular}
    \end{table}

To follow up the significant 2-way interactions in Model 2, we conducted post-hoc pairwise comparisons with Tukey's correction. For the effect of trial within each condition, there was no improvement in the screen condition in time to add the object across the 3 trials, whereas for the VR condition, trial 3 was significantly faster than trial 2 and trial 1 (see \cref{tab:posthoc_add_trialxcondition}). 

            \begin{table}[h!]
            \caption{Simple Effects of Trial by Condition on Time to Add Object}
            \centering
            \footnotesize
            \label{tab:posthoc_add_trialxcondition}
            \begin{tabular}{lllrrrr}
              \toprule
               \textbf{VR} &  \textit{b} & \textit{se} & \textit{95\% CIs} &\textit{t} & \textit{p} \\ 
              \midrule
               Trial 1 - Trial 2 & 0.04 & 0.06  & [-0.11, 0.63] & 0.63 & .802 \\
               Trial 1 - Trial 3 & 0.26 & 0.07  & [0.11, 0.42] & 4.03 & $<.001$ \\  
               Trial 2 - Trial 3 & 0.22 & 0.07 & [0.07, 0.38] & 3.39 & .003 \\  \midrule
                 \multicolumn{4}{l}{\textbf{Screen}} \\ \midrule
               Trial 1 - Trial 2 & 0.13 & 0.06  & [-0.02, 0.28] & 2.02 & .120 \\
               Trial 1 - Trial 3 & 0.09 & 0.07  & [-0.07, 0.24] & 1.31 & .393 \\  
               Trial 2 - Trial 3 & -0.05 & 0.06 & [-0.20, 0.11] & -0.69 & .769 \\  
               \bottomrule
            \end{tabular}
            \end{table}

For the effect of condition within each order, when participants started with the screen interface, they took significantly less time to add the objects in the VR condition, and vice versa for when they started with the VR, see \cref{tab:posthoc_add_conditionxorder}. This suggests that participants were always faster in the second interface they used. However, as the 3-way interaction model did not better explain the data, we can still be confident in interpreting the 2-way interaction between condition and trial.  

\begin{table}[h!]
    \caption{Simple Effects of Condition by Order on Time to Add Object}
    \centering
    \footnotesize
    \label{tab:posthoc_add_conditionxorder}
    \begin{tabular}{lllrrrr}
        \toprule
         &  \textit{b} & \textit{se} & \textit{95\% CIs} &\textit{t} & \textit{p} \\ 
         \midrule
         \textbf{VR First} & & & & & \\
            VR - Screen & 0.16 & 0.05 & [0.26, 0.05] & 2.99 & .003 \\  \midrule
        \textbf{Screen First} & & & & & \\
            VR - Screen & -0.2 & 0.05 & [-0.10, -0.31] & 3.86 & $<.001$ \\  
        \bottomrule
    \end{tabular}
    
    \end{table}

\subsubsection{Time to Place Object}
\begin{figure}[t]
    \centering
    \includegraphics[width=0.5\textwidth]{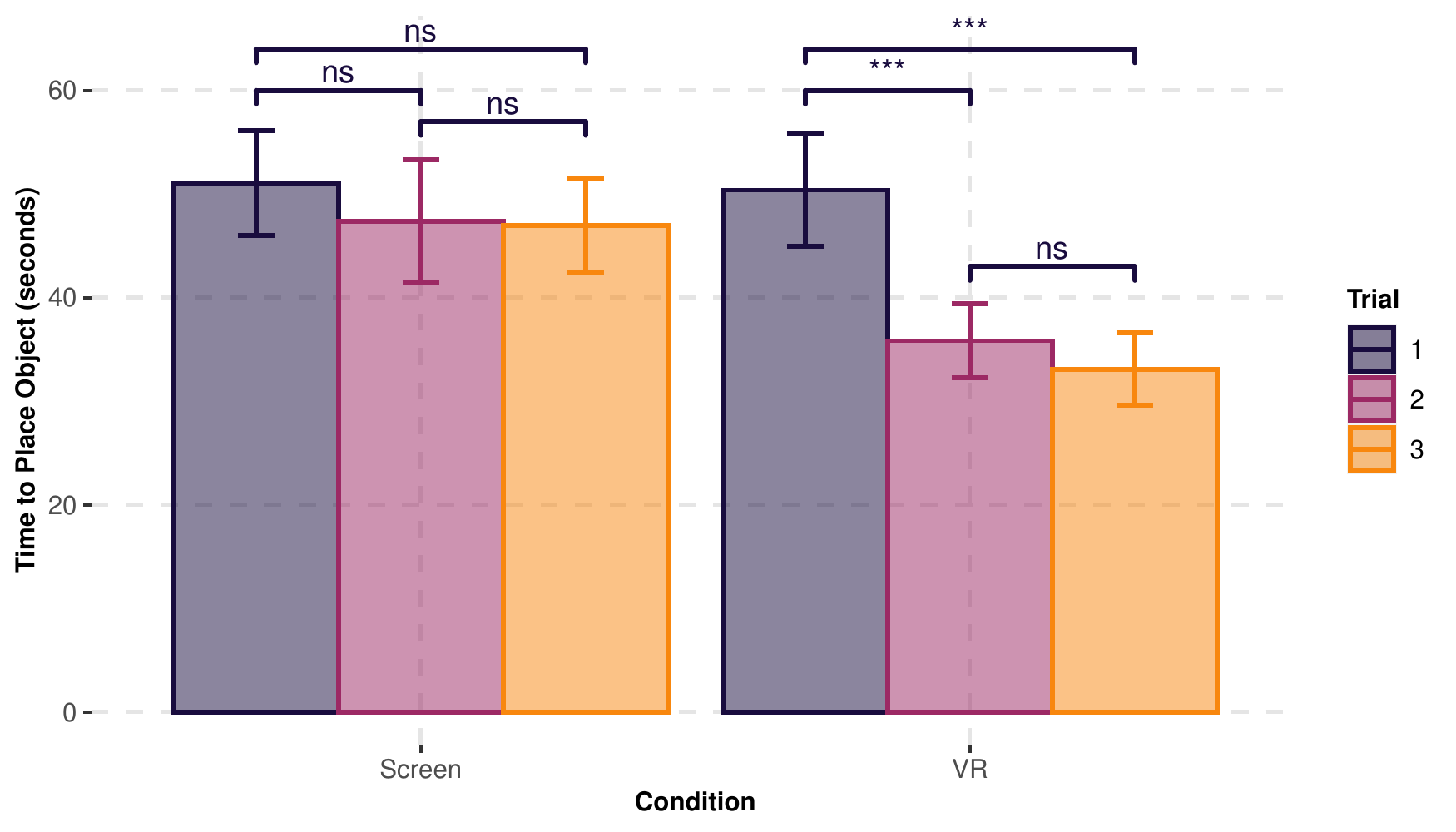}
    \caption{Interaction between Condition and Trial for Time to Place Object.}
    \label{fig:place_interact}
\end{figure}

The model for the time taken from when participants added the missing object until they executed the movement on the virtual robot (Model 4) showed a significant fixed effect of condition, see \cref{tab:place_fixef}. Similar to before, we fitted models with the 2- and 3-way interaction terms between condition, trial, and order (Model 5, Model 6), see \cref{tab:place_interact}. Model 5 provided the best fit for the data, with significant 2-way interactions between condition and trial (see \cref{fig:place_interact}), and condition and order.

 \begin{table}[b]
    \centering
    \scriptsize
    \caption{Fixed Effects for Time to Place Object}
    \label{tab:place_fixef}
    \begin{tabular}[width=\columnwidth]{lllllll}
        \toprule
        \multirow{2}{*}{\textbf{Variable}} & \multicolumn{6}{c}{\textbf{Model 4}} \\ \cmidrule(lr){2-7} 
                 & \textit{b} & \textit{se} & \textit{t} & \textit{F} & \textit{p} & \textit{95\% CIs} \\ \midrule
                (Intercept) & 2.13 & 0.42 & 59.09 & & $<.001$ & [2.06, 2.20] \\ 
                  Condition & -0.13 & 0.02 & -5.17 & & $<.001$ & [-0.18, -0.08] \\ 
                  Order & 0.01 & 0.04 & 0.21 & & .831 & [-0.07, 0.09] \\
                  FAM-VAM & -0.05 & 0.03 & -1.85 & & .058 & [-0.11, 0.00]  \\
                  FAM-Robot & 0.01 & 0.02 & 0.32 & & .740 & [-0.03, 0.05]  \\
                  Trial &  &  &  & 13.44 & .001  \\ \midrule
                  AIC & 34.15 & & & & \\ \bottomrule
    \end{tabular}
    \end{table}

\begin{table}[b]
    \centering
    \footnotesize
    \caption{Interaction Effects for Time to Place Object}
    \label{tab:place_interact}
    \begin{tabular}[width=\columnwidth]{lllll}
    \toprule
                          & \multicolumn{2}{c}{\textbf{Model 5}} &  \multicolumn{2}{c}{\textbf{Model 6}}  \\ \cmidrule(lr){2-3} \cmidrule(lr){4-5} 
                          & \textit{F}        & \textit{p} &  \textit{F}       & \textit{p} \\ \midrule
      Condition x Trial   & 8.03 & .018 & & \\
      Condition x Order   &  7.00 & .008   & & \\
      Order x Trial       &  0.47 & .789  & & \\
      Condition x Trial x Order &  &  & 17.02 & .017 \\ \addlinespace \midrule
      AIC                  & 28.86 & & 31.12 & \\ \bottomrule
    \end{tabular}
    \end{table}

Post-hoc pairwise comparisons with Tukey's correction of the difference in time between each trial within each condition showed the same pattern as the time taken to add object; with no difference in any of the screen trials, but a significant improvement from trial 1 to trial 3, and from trial 1 to trial 2 in the VR condition, see \cref{tab:posthoc_place_trialxcondition}. 

    \begin{table}[t]
    \caption{Simple Effects of Trial by Condition on Time to Place Object}
    \centering
    \footnotesize
    \label{tab:posthoc_place_trialxcondition}
    \begin{adjustbox}{max width=\columnwidth}
    \begin{tabular}{lllrrrr}
      \toprule
     \textbf{VR}  &  \textit{b} & \textit{se} & \textit{95\% CIs} &\textit{t} & \textit{p} \\ 
      \midrule
       Trial 1 - Trial 2 & 0.17 & 0.04  & [0.07, 0.27] & 4.11 & $<.001$ \\
       Trial 1 - Trial 3 & 0.17 & 0.04  & [0.07, 0.27] & 3.91 & $<.001$ \\  
       Trial 2 - Trial 3 & -0.01 & 0.04 & [-0.11, 0.10] & -0.15 & .987 \\  \midrule
         \multicolumn{4}{l}{\textbf{Screen}} \\ \midrule
       Trial 1 - Trial 2 & 0.03 & 0.04  & [-0.07, 0.13] & 0.62 & .811 \\
       Trial 1 - Trial 3 & 0.02 & 0.04  & [-0.08, 0.13] & 0.56 & .844 \\  
       Trial 2 - Trial 3 & 0.00 & 0.04 & [-0.10, 0.10] & -0.05 & .998 \\  
       \bottomrule
    \end{tabular}
    \end{adjustbox}
    \end{table}

When participants started with the screen, they were faster at placing the object in VR, whereas there was no difference when starting with VR, \cref{tab:posthoc_place_conditionxorder}.

    \begin{table}[b]
        \caption{Simple Effects of Condition by Order on Time to Place Object}
        \centering
        \footnotesize
        \label{tab:posthoc_place_conditionxorder}
        \begin{tabular}{lllrrrr}
            \toprule
             &  \textit{b} & \textit{se} & \textit{95\% CIs} &\textit{t} & \textit{p} \\ 
             \midrule
         \textbf{VR First} & & & & & \\
                VR - Screen & -0.06 & 0.03 & [-0.00, 0.13] & 1.92 & .056 \\  \midrule
         \textbf{Screen First} & & & & & \\
                VR - Screen & -0.19 & 0.03 & [-0.12, -0.26] & 5.56 & $<.001$ \\  
            \bottomrule
        \end{tabular}
        \end{table}

\subsubsection{Final Object Distance}
For the final distance of the object, we performed a generalized linear mixed model with a Gamma distribution and inverse link. Fixed effects included condition, order, trial, and (mean-centered) familiarity with VAM systems and robots. As a random effect we again nested trial within participant ID, see \cref{tab:nmissing_fixef}. This model only showed an effect of familiarity with VAM systems; such that greater familiarity led to better object placement. No other fixed effects were significant (AIC = -1740.1). Including the 2 and 3-way interaction terms between condition, trial, and order did not lead to a better model fit (AIC = -1739.9 and AIC = -1736.3 for the two- and three-way interaction models, respectively). 

\subsubsection{Number of Times Missing Object Added}
\begin{figure}
    \centering
    \includegraphics[width=0.5\textwidth]{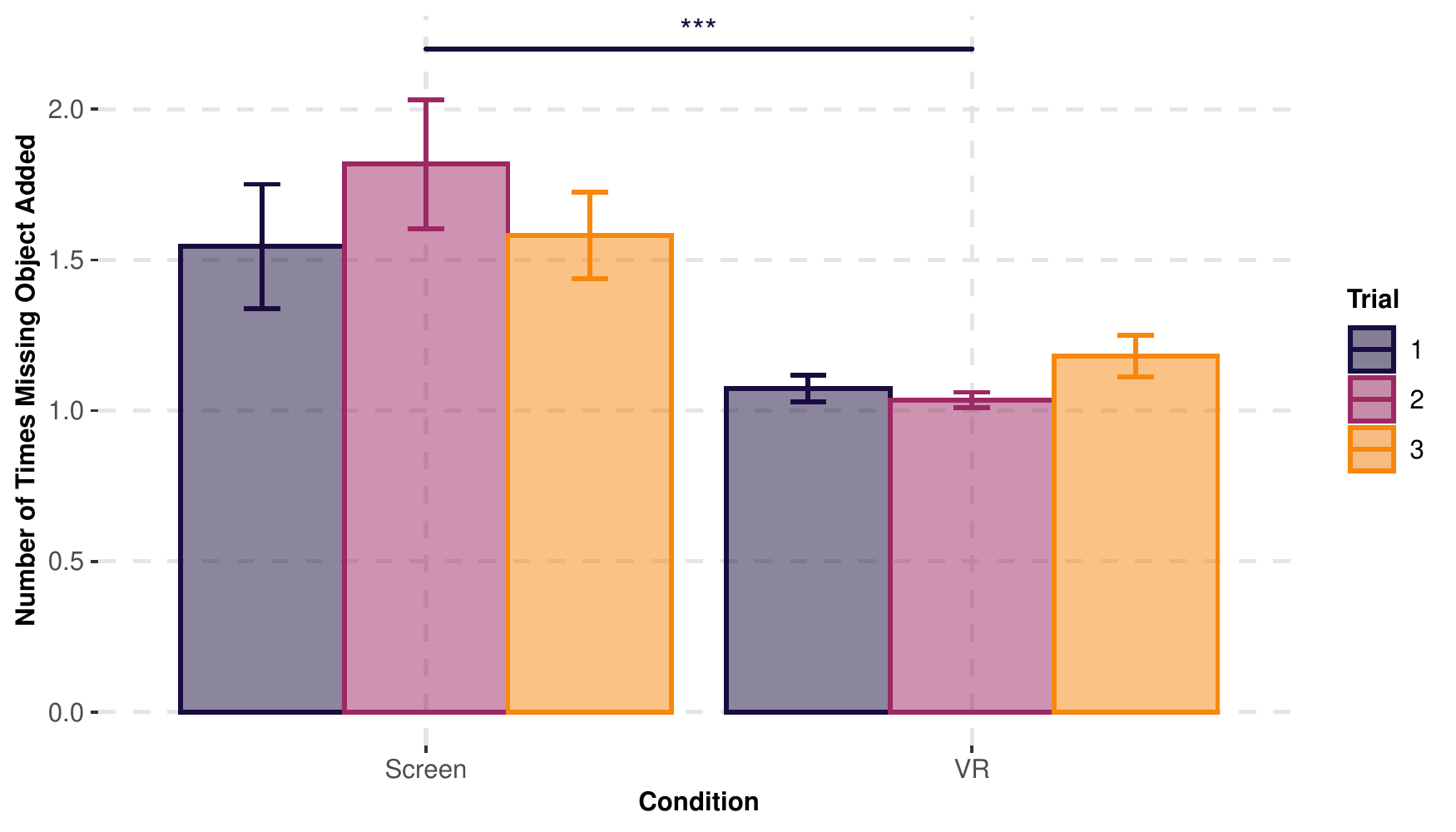}
    \caption{Interaction between Condition and Trial for number of times Missing Object Added.}
    \label{fig:missing_interact}
\end{figure}

We counted the number of times participants added the missing object to the environment and performed a generalized linear mixed model with a Poisson distribution and log link. We specified fixed effects of condition, order, trial, and (mean-centered) familiarity with VAM systems and robots and nested trial within participant ID as a random effect, see \cref{tab:nmissing_fixef}. We found a significant fixed effect of condition, such that the missing object was added significantly more in the screen condition compared to the VR, see \cref{fig:missing_interact}. Comparing models with the two and three-way interactions of condition, trial, and order with the original fixed-effects model did not improve model fit (AIC = 876.46, AIC = 883.94, AIC = 887.85). 

\begin{table}[b]
    \centering
    \footnotesize
    \caption{Fixed Effects for Number of Times Missing Object Added}
    \label{tab:nmissing_fixef}
    \begin{adjustbox}{max width=\columnwidth}
    \begin{tabular}{lllllll}
      \toprule
    \multirow{2}{*}{\textbf{Variable}} & \multicolumn{6}{c}{\textbf{Model 7}} \\ \cmidrule(lr){2-7} 
     & \textit{b} & \textit{se} & \textit{z} & \textit{F} & \textit{p} & \textit{95\% CIs} \\ \midrule
    (Intercept) & 0.60 & 0.18 & 3.41 & & $<.001$ & [0.25, 0.94]  \\ 
      Condition & -0.41 & 0.10 & -4.25 & & $<.001$ & [-0.60, -0.22]  \\ 
      Order & 0.07 & 0.11 & 0.66 & & .508 & [-0.15, 0.30]  \\
      FAM-VAM & -0.14 & 0.08 & -1.68 & & .093 & [-0.30, 0.02] \\
      FAM-Robot & 0.03 & 0.06 & 0.48 & & .632 & [-0.09, 0.14]  \\
      Trial &  &  &  & 0.658 & .720 \\ \bottomrule
    \end{tabular}
    \end{adjustbox}
    \end{table}

\subsection{Subjective responses}
The enjoyment scale was reliable for both the VR ($\alpha = 0.94$) and screen ($\alpha = 0.90 $) conditions. However, the residuals were highly positively skewed and failed to achieve normality after transformation. Consequently, we performed a Friedman test specifying enjoyment as the outcome and condition as the within-groups independent variable, which was non-significant, $\chi(1) = 0.32, p = .572$. The high positive skew could imply a ceiling effect, that is, most participants enjoyed the task overall, irrespective of condition. 

Although all of the presence subscales were reliable when applied to the VR condition (all $\alpha 's > 0.61 $), the same was not true for the screen condition (all $\alpha 's < 0.65 $). Consequently, we did not perform further analyses on the presence scale. 

\begin{figure}
    \centering
    \includegraphics[width=0.5\textwidth]{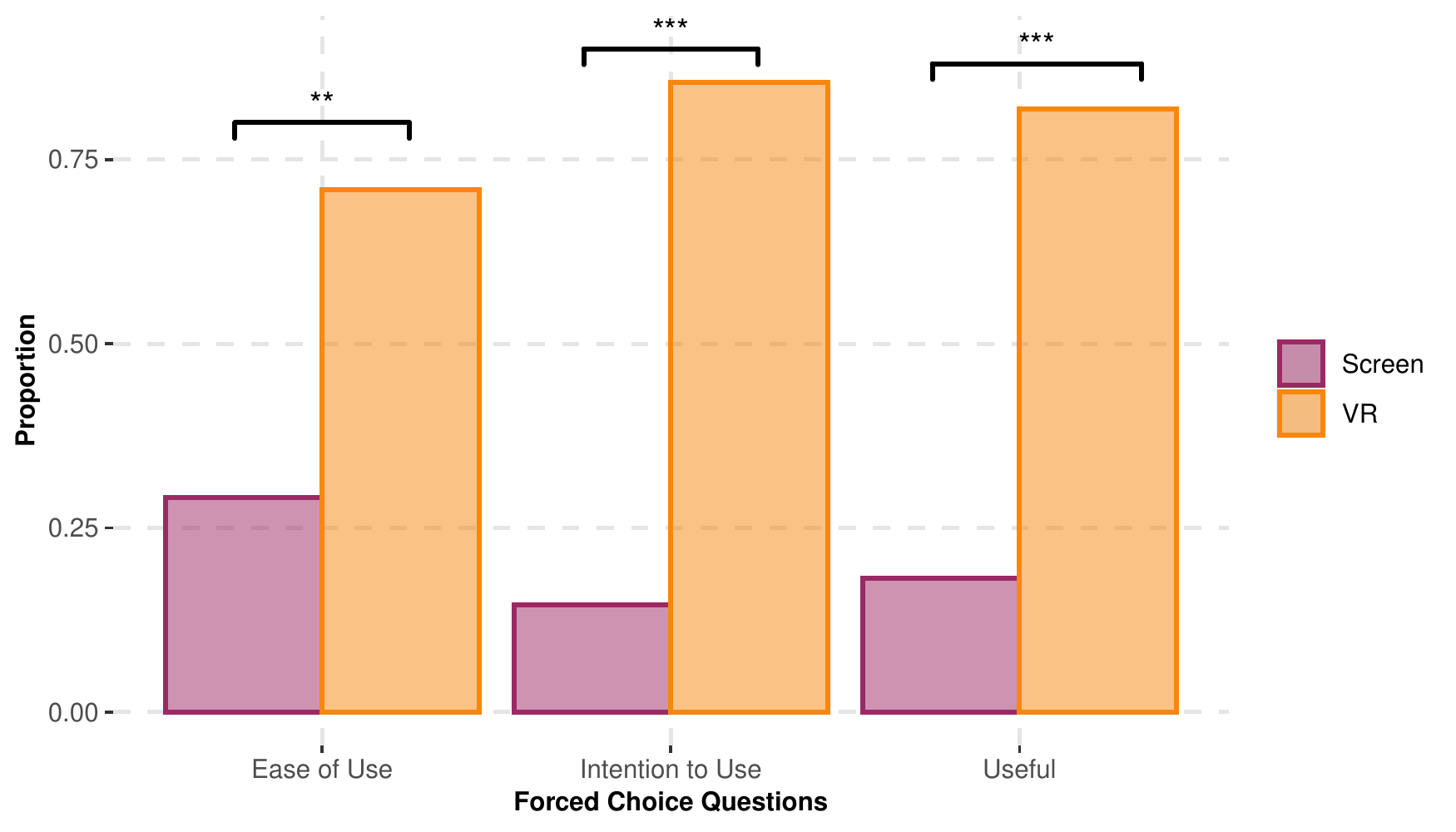}
    \caption{Proportion of Participants who selected the VR versus the Screen Interface for Ease of Use, Usefulness, and Intention to Use.}
    \label{fig:tam_prop}
\end{figure}

We compared participants' responses to the three forced-choice questions using one-sample t-tests of proportions. All 3 questions showed a strong preference for the VR over the Screen interface ($mu = 0.5$, all $t's > 3.17$, all $p's < .001$), see \cref{fig:tam_prop}.

\subsubsection{Qualitative Observations}
The findings from the forced-choice questions were supported by the open-ended responses. In particular, three main themes could be identified. First, that the VR condition was more fun (``VR was exciting and superfun'', ``the VR was more fun to use''). Second, it was more intuitive and/or immersive (``VR made me feel more in control of my own actions as well as the robots'', ``the VR felt more like real-life''), and third, it was easier to perform the task (``VR had more accurate object placement'', ``overall [VR] allowed more precise movements'')


Some participants also expressed that although they initially expected the screen and keyboard to be easier, in the end, they preferred the VR condition; ``I would think it would be easier to use the keyboard and mouse at first [...], but after the first trial the VR was way more practical and intuitive.''


Other participants reported that even if they found some aspects difficult in the VR condition, they still enjoyed the experience more than the screen and keyboard; ``The screen was easier to use, but the VR was more informative and fun'',``VR makes it more realistic and entertaining even though is less practical''.






Some participants also expressed concern about using VR for an extended period of time; ``I found it easier to use the VR tool but using it during more than half an hour could make me more uncomfortable'', ``VR was easier in a way but I maybe would get a headache when doing it longer''.

\section{Discussion}
\label{sec:diss}
We first hypothesised that participants would select the VR interface as more useful, easier to use, and would prefer to interact with the VR again over the screen interface (H1a-H1c). This hypothesis was supported, with participants indicating a strong preference for VR across all 3 questions.

Our second hypothesis was that task performance would be better in VR than with the screen, which was only partially supported. We found no difference between the screen and VR with respect to how accurate participants were at placing the object (H2a). The final distance was calculated from the center of the object to the center of the bounding box of the detected object. However, this approach is imperfect since the bounding box does not accurately represent the shape of the object. Furthermore, it does not take into account any object rotation, which can also affect the final placement accuracy. Consequently, findings related to the final placement of the object should be interpreted with caution. 


However, we did find an interaction between condition and trial number, supporting H2b. For both the time taken to add the missing object to the virtual environment and the time taken to place the object, participants showed a significant improvement across trials in the VR condition, but not in the screen condition. 


Finally, we hypothesised that participants would enjoy the VR system more (H3a) and report higher ratings of presence (H3b) compared to the screen interface. Although the self-report questionnaires could not be analysed quantitatively, qualitative responses from the participants suggest that the VR was perceived as more enjoyable and immersive. 




\subsection{Limitations and Future Work}
The presence scale from \cite{witmer1998measuring} was initially designed with VR systems in mind. The decision to adapt items to the screen and keyboard context in this study was made based on a lack of existing measures for comparing VR and screen interfaces. Very few studies in HRI have conducted comparisons between VR and screen and keyboard interfaces. One direction for future work is therefore to continue developing measures that can be used to assess attitudes toward both screen and VR interfaces.

Although participants were only required to add and move the object that failed to be detected in each trial, we observed that many participants also chose to move the other objects present in the virtual environment to improve their placement. The virtual world in this framework is only a representation of the real world. Thus, it may have been difficult to determine what exactly constituted a perceptual error, as the objects do not look exactly alike in the real and virtual worlds.  

Additionally, many participants attempted to rotate the objects to better maximise the position accuracy. In our framework, objects are detected only based on bounding boxes, thus the exact rotation of the object is not necessary for the robot to be able to pick up and move an object. Consequently, there may have been a disconnect between the actual functioning of the system, and how users expected the system to behave. Future iterations of the framework will focus on improving the transparency of the robot's behaviour and improving the user interface to better align with user's expectations (i.e., allowing for object rotation).

One potential application of our framework is improving the robot's perception module by adding missing objects in the VR environment. 
The deep learning model, responsible for the robot's perception, could be retrained using collected and additional (from other sources or even generated) data. By using this approach, it is possible to iteratively refine and improve the robot perception module over time, ultimately resulting in a more accurate and robust system.



\section{Conclusion}
This study aimed to test how well a virtual reality interface for correcting robot errors performs against a baseline comparison of a screen and keyboard interface. Participants improved much faster over the trials with the VR, compared to the screen and keyboard, which did not show any significant improvement over time. Participants also consistently chose the VR over the screen interface as more useful, easier to use, and their preferred interface for interacting with again. Furthermore, VR showed a steeper learning curve, suggesting that users' performance improves faster in VR than with the screen interface.  These findings provide support for the use of VR as a means of correcting robot errors. 



%
\section{Acknowledgments}
This work was partially supported by the Wallenberg AI, Autonomous Systems and Software Program (WASP) funded by the Knut and Alice Wallenberg Foundation and the Vinnova Competence Center for Trustworthy Edge Computing Systems and Applications. We also thank Lucas Morillo for comments on the initial manuscript.

\bibliographystyle{ieeetr}
\bibliography{ref}
\end{document}